\documentclass[a4paper]{svproc}
\usepackage{url}

\usepackage{amsmath, amssymb, amsfonts, mathtools}
\usepackage{xcolor}
\usepackage{bm}
\usepackage{scalerel}
\usepackage{multirow}
\usepackage{makecell}
\usepackage{booktabs}
\usepackage{adjustbox}
\usepackage{subcaption}

\DeclarePairedDelimiterX{\norm}[1]{\lVert}{\rVert}{#1}

\usepackage[
backend=biber,
style=ieee,
]{biblatex}
\usepackage{wrapfig}

\usepackage[para, online, flushleft]{threeparttable}
\usepackage{array, booktabs, makecell}
\usepackage{bm}

\usepackage[inline]{enumitem}
\usepackage{booktabs}
\usepackage{relsize}

\usepackage{tabularx}
\usepackage{hyperref}

\begin{document}
\mainmatter              
%
\title{Can Large Language Models Solve Robot Routing?}
\titlerunning{Can Large Language Models Solve Robot Routing?}  
%
\author{Zhehui Huang \and Guangyao Shi \and Gaurav S. Sukhatme}
\authorrunning{Zhehui Huang et al.} 
%
\tocauthor{Zhehui Huang, Guangyao Shi Gaurav S. Sukhatme}
\institute{Department of Computer Science\\University of Southern California, Los Angeles, CA, 90089, USA\\
\email{zhehuihu@usc.edu, shig@usc.edu, gaurav@usc.edu
}}

\maketitle              

\begin{abstract}

Routing problems are common in mobile robotics, encompassing tasks such as inspection, surveillance, and coverage. Depending on the objective and constraints, these problems often reduce to variants of the Traveling Salesman Problem (TSP), with solutions traditionally derived by translating high-level objectives into an optimization formulation and using modern solvers to arrive at a solution. 
Here, we explore the potential of Large Language Models (LLMs) to replace the entire pipeline from tasks described in natural language to the generation of robot routes. 
We systematically investigate the performance of LLMs in robot routing by constructing a dataset with 80 unique robot routing problems across 8 variants in both single and multi-robot settings. 
We evaluate LLMs through three frameworks — single attempt, self-debugging, and self-debugging with self-verification — and various contexts, including mathematical formulations, pseudo-code, and related research papers. 
Our findings reveal that both self-debugging and self-verification enhance success rates without significantly lowering the optimality gap. 
We observe context-sensitive behavior - providing mathematical formulations as context decreases the optimality gap but significantly decreases success rates, and providing pseudo-code and related research papers as context does not consistently improve success rates or decrease the optimality gap. 
We identify key challenges and propose future directions to enhance LLM performance in solving robot routing problems. Our source code is available on the project website
~\href{https://sites.google.com/view/words-to-routes/}{https://sites.google.com/view/words-to-routes/}.

\end{abstract}
\section{Introduction}

\begin{figure}[ht]
\centering
\includegraphics[scale=0.78]{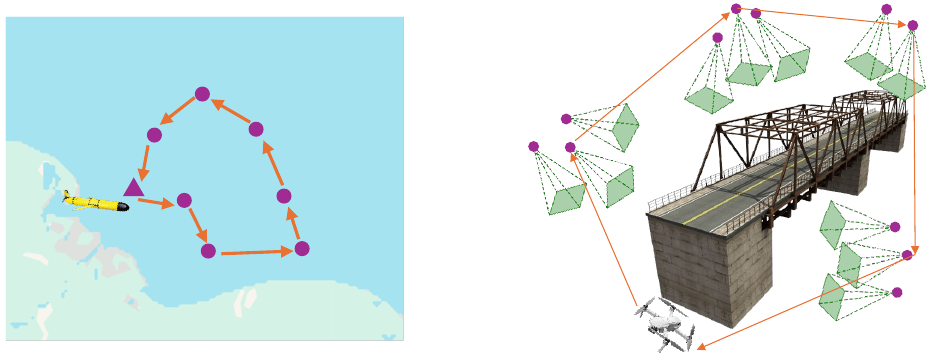}
\vspace{-0.1in}
\caption{
Two illustrative examples of robot routing problems. (left) a marine robot needs to visit a set of candidate locations to take ocean samples, and the operator wants the robot to finish the sampling process as fast as possible - this corresponds to a Traveling Salesman Problem (TSP); (right) an aerial robot with cameras needs to visit a subset of all the candidate viewpoints to ensure sufficient coverage of the bridge, which corresponds to the generalized TSP.   
}
\label{fig:overview}
\vspace{-0.26in}
\end{figure}

Routing problems are ubiquitous in robotics, in domains spanning inspection~\cite{hollinger2012uncertainty}, surveillance~\cite{stump2011multi}, sampling~\cite{ma2018data}, monitoring~\cite{ma2017informative, shi2023robust}, search and rescue~\cite{morin2023ant} and coverage~\cite{gasparri2008framework, karapetyan2023ag} 
 (Fig. \ref{fig:overview}). 
The fundamental question is to plan a path that a robot (or multiple paths for multiple robots) must follow to execute a task - usually framed as a sequence of locations that a robot must visit. Depending on the precise objective(s) and the associated constraint(s), such problems often correspond to variants of the Traveling Salesman Problem (TSP). 
Historically, the high-level objective the end-user has in mind ("inspect a bridge") is manually translated into a mathematical objective and constraints by an expert (e.g., "traverse all spatial viewpoints from set \textit{S} and minimize the total travel time \textit{T}"). 
The resulting optimization problem is then exactly or approximately solved using a solver which produces the plan for the robot(s). This is then handed off to a low-level controller to execute.

The emergence of Large Language Models (LLMs) leads to two natural questions: 1. Can an LLM replace the manual translation of end-user natural language-specified desiderata to a mathematical formulation of an optimization problem; and 2. Can one go one step further and replace the entire pipeline - translation from a natural language specification to an optimization problem formulation followed by the generation of a plan - with an LLM? The first question - essentially a hybrid approach - has received some attention in the literature in planning~\cite{liu2023llm+} and navigation~\cite{shah2023lm}. The second question - essentially an end-to-end approach - has also been explored to some extent~\cite{chen2024solving, kambhampati2024llms} largely with negative results. 

Is the end-to-end route really a dead-end or could it be of some use in robot routing? We systematically investigate this question by constructing and experimenting with a dataset composed of 80 unique problems across 8 widely studied variants of routing problems (4 for a single robot and 4 for multiple robots). 

\begin{figure}[ht!]
\centering
\includegraphics[scale=0.26]{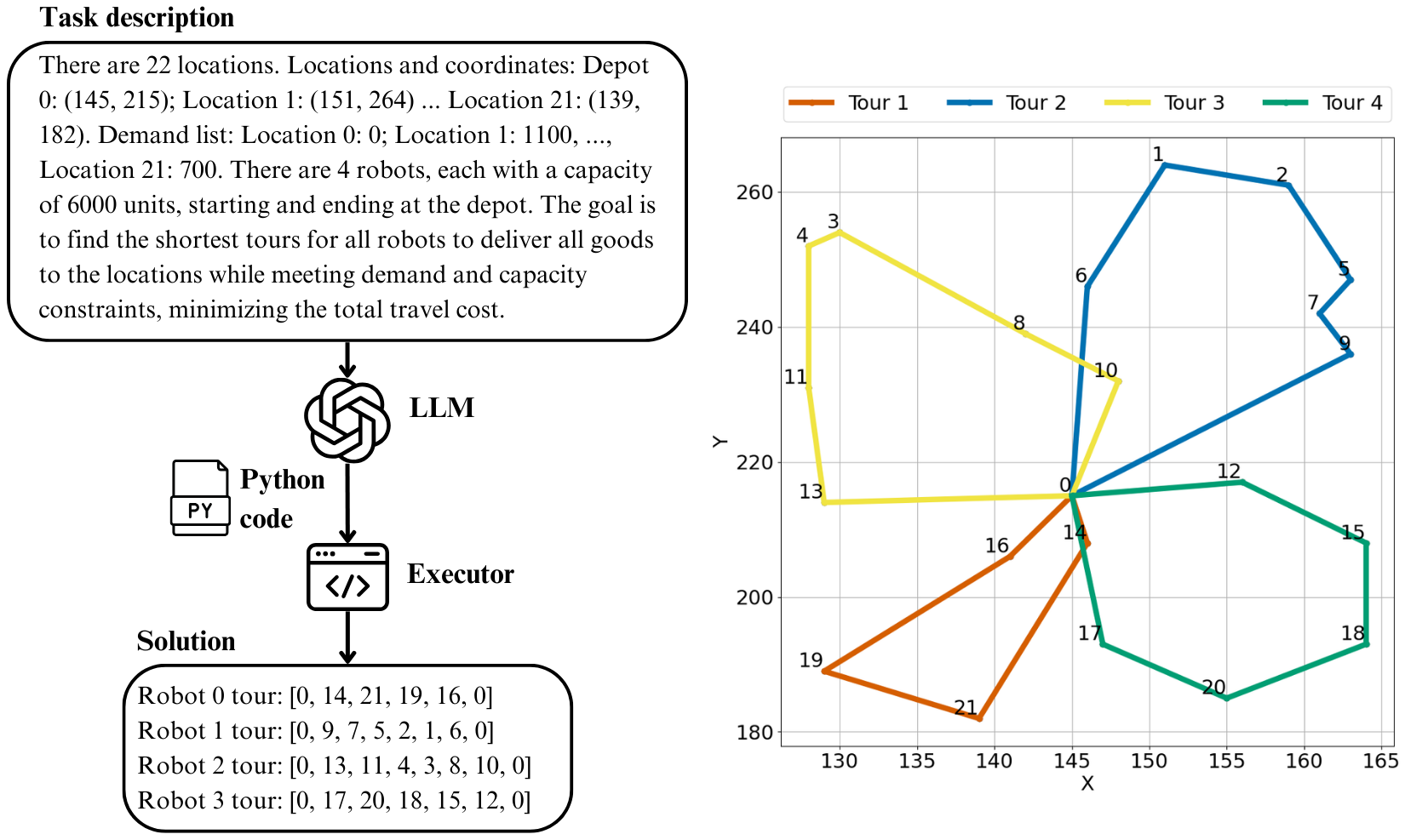}
\caption{An illustrative scenario based on the Capacitated Vehicle Routing Problem (CVRP). Left: an overview of acquiring solutions given task descriptions in a zero-shot manner. Right: solution visualization.}
\label{fig:overview_v2}
\vspace{-0.24in}
\end{figure}

After constructing the dataset, we evaluate the performance of LLMs along three dimensions. 
First, we design three different frameworks - \texttt{single attempt}, \\ \texttt{self-debugging}, and \texttt{self-debugging with self-verification}. We provide an LLM only with task descriptions and measure its effectiveness in these three frameworks for solving routing problems. 
As an example, Fig.~\ref{fig:overview_v2} shows an overview of the \texttt{single-attempt} framework. 

Second, we provide context to the LLM, including the exact \textsc{mathematical formulation} of the problem, the \textsc{pseudo-code} of a heuristic algorithm, and a research \textsc{paper} related to the problem. 

Third, we evaluate the performance of different LLMs on robot routing. 

The main contributions of our paper are:

\begin{itemize}
    \item We construct a dataset with 80 unique robot routing problems across 8 variants, including 4 variants for single robot routing and 4 variants for multi-robot routing, to serve as a benchmark for LLMs in robot routing. This is the first dataset of its kind.
    \item Using the dataset, we perform a comprehensive study to evaluate the performance of LLMs in solving robot routing problems using different frameworks and different contexts.
    \item We discuss what impedes LLMs from generating better solutions for robot routing and propose potential directions for future work to improve their performance.
\end{itemize}

\section{Related Work}
\subsection{LLMs for Routing Problems}
There is very limited work that specifically uses LLMs to solve routing problems. However, there are some works that focus on integrating LLMs into the optimization algorithms that use the classical single TSP as a test case~\cite{yang2023large, liu2023large, liu2023algorithm, ye2024reevo} or those that use the CVRP as a test case~\cite{ye2024reevo, huang2024multimodal}. 
Yang et al.~\cite{yang2023large} use LLMs as optimizers and iteratively generate new solutions given task descriptions and previous solution-score pairs. 
Liu et al.~\cite{liu2023large}, Liu et al.~\cite{liu2023algorithm}, and Ye et al.~\cite{ye2024reevo} focus on combining LLMs with evolutionary algorithms. 
Specifically, Liu et al.~\cite{liu2023large} ask LLMs to select parent solutions and perform crossover and mutation to generate offspring solutions. 
Liu et al.~\cite{liu2023algorithm}, instead of asking LLMs to operate on solutions, ask LLMs to operate on heuristic algorithms. 
Ye et al.~\cite{ye2024reevo} share the same idea as Liu et al.~\cite{liu2023algorithm}, but introduce short-term reflection and long-term reflection mechanisms to better use history heuristics. 
Huang et al.~\cite{huang2024multimodal} use multimodal LLMs to solve optimization problems and rely on XML text as additional input to solve the problem. 

Unlike previous works that either extract heuristics from LLMs or use LLMs to solve problems based on well-crafted prompts, 
our work focuses on directly generating solutions from LLMs based on natural language task descriptions.

\subsection{LLMs for Code Generation and Code Repair}

LLMs have shown remarkable progress in text-to-code generation without specialized finetuning~\cite{nijkamp2022codegen}.
However, the generated code might be incorrect. 
Many works have been proposed to solve this problem. 
We split them into two groups. 
The first group of works~\cite{wang2022self, zhang2023coder} aims to generate multiple solutions initially followed by selecting the best one based on varying criteria for determining the 'best' solution. 
The second group of works focuses on generating an initial solution and then using LLMs to regenerate solutions based on feedback, such as execution results. This process of solution regeneration is known as self-reflection or self-refinement. There are various types of self-reflection, depending on the nature of the feedback~\cite{madaan2024self, zhang2024context}. 

In this work, inspired by the idea that high-quality feedback can enhance the performance of LLMs~\cite{chen2023teaching}, we utilize the fact that constraints in routing problems are naturally described within the task. For example, a robot must visit all locations exactly once. To amplify the benefits of these constraints, we first instruct an LLM to extract constraints from natural language task descriptions. Then, we ask the LLM to generate unit tests based on these constraints.

\section{Dataset}\label{sec:dataset}
To evaluate the capability of LLMs to solve routing problems given natural language task descriptions, we construct a dataset that contains four variants of single-robot and four variants of multi-robot routing. 
For each single robot-based variant, we design cases with $10$, $15$, and $20$ locations to visit. 
For each number of locations, we randomly generate five instances with positions as $x-y$ coordinates within [0, 100]. 
Each multiple robot-based variant is evaluated based on five instances selected from CVRPLIB\footnote{https://vrp.atd-lab.inf.puc-rio.br/index.php/en/}. 
Specifically, these five instances are 
P-n16-k8,
P-n19-k2,
P-n21-k2,
E-n22-k4,
P-n23-k8.
Problems that start with P are from~\cite{augerat1995approche}, and problems that start with E are from~\cite{christofides1969algorithm}.
The value after $n$ denotes the number of locations, and the number after $k$ denotes the number of robots.
For all instances, we implement solutions based on Gurobi~\cite{gurobi} to obtain optimal results. 
The definitions of all variants are given as follows, and the corresponding IP formulations are available on our website\footnote{\href{https://sites.google.com/view/words-to-routes/}{https://sites.google.com/view/words-to-routes/}}. 
\subsection{Single Robot}
\textbf{Traveling Salesman Problem (TSP)}\cite{hoffman2013traveling}: 
Given $n$ locations and the distance between each pair of locations, find the shortest possible robot tour that visits each location exactly once and returns to the depot.

\noindent  \textbf{Bottleneck Traveling Salesman Problem (BTSP)}\cite{kabadi2007bottleneck}: 
Given $n$ locations and the distance between each pair of locations, find a robot tour that minimizes the maximum distance between any two consecutive locations in the tour and returns to the depot.

\noindent  \textbf{$k$-Traveling Salesman Problem ($k$-TSP)}\cite{pandiri2020two}:
Given $n$ locations, the distance between each pair of locations, and a fixed value $k$ such that $1 < k \leq n$, the $k$-TSP problem seeks a robot tour of minimum length that starts and ends at the depot, visiting exactly $k$ locations (including the depot) out of $n$ locations.

\noindent  \textbf{Generalized Traveling Salesman Problem (GTSP)} \cite{pop2024comprehensive}: Given groups of locations, find the shortest robot tour that visits exactly one location from each cluster and returns to the depot.

\subsection{Multiple Robots}
\noindent  \textbf{Multiple Traveling Salesman Problem ($m$-TSP)}~\cite{bektas2006multiple}: Given $n$ locations and $m$ robots located at a single depot, find tours for all robots, which start and end at the depot, such that each location is visited exactly once and the sum of the total cost of visiting all locations is minimized.

\noindent \textbf{MinMax Multiple Traveling Salesman Problem (MinMax $m$-TSP)}~\cite{matsuura2014solving}: Given $n$ locations and $m$ robots located at a single depot location, find tours for all robots, which start and end at the depot, such that each location is visited exactly once and the maximum distance traveled by any robot is minimized.

\noindent \textbf{Multi-Depot Multi-Traveling Salesman Problem (MD $m$-TSP)}~\cite{kara2006integer}: 

\noindent Given $n$ locations and $m$ robots located at $d \leq m$ depots, find tours for all robots. 
Each robot starts and ends at its own depot, ensuring that each location is visited exactly once, and the total cost of visiting all locations is minimized.

\noindent  \textbf{Capacitated Vehicle Routing Problem (CVRP)}~\cite{borcinova2017two}: Given $n$ customers, each with a specific demand, the objective is to determine a set of tours for a fleet of capacitated robots based on a single depot. The goal is to minimize the total cost of the tours, ensuring that each tour begins and ends at the depot, each customer is visited exactly once, and the total demand served by each tour does not exceed the robot's capacity.

\section{Method}\label{sec:method}
To comprehensively assess the capability of using LLMs for robot routing problems based solely on natural language task descriptions, we conduct evaluations along three main dimensions: examining the effectiveness of various LLM frameworks, analyzing the impact of incorporating different contexts, and evaluating the performance of different LLMs.

\subsection{LLM Framework Design}
We start with a framework that only asks LLMs to generate a solution once as a baseline. We refer to this framework as \texttt{single attempt} in the following. High-quality feedback (i.e., execution error) shows promising performance in increasing the performance of LLMs~\cite{chen2023teaching, kim2024language, shinn2024reflexion}. 
Inspired by this, we designed the second framework building on the baseline framework. The second framework allows LLMs to regenerate solutions given execution errors if generated solutions fail in execution. 
In the following, we refer to this framework as \texttt{self-debugging}. 

\begin{figure*}[ht]
\centering
\vspace{-0.2in}
\includegraphics[width=0.95\linewidth]{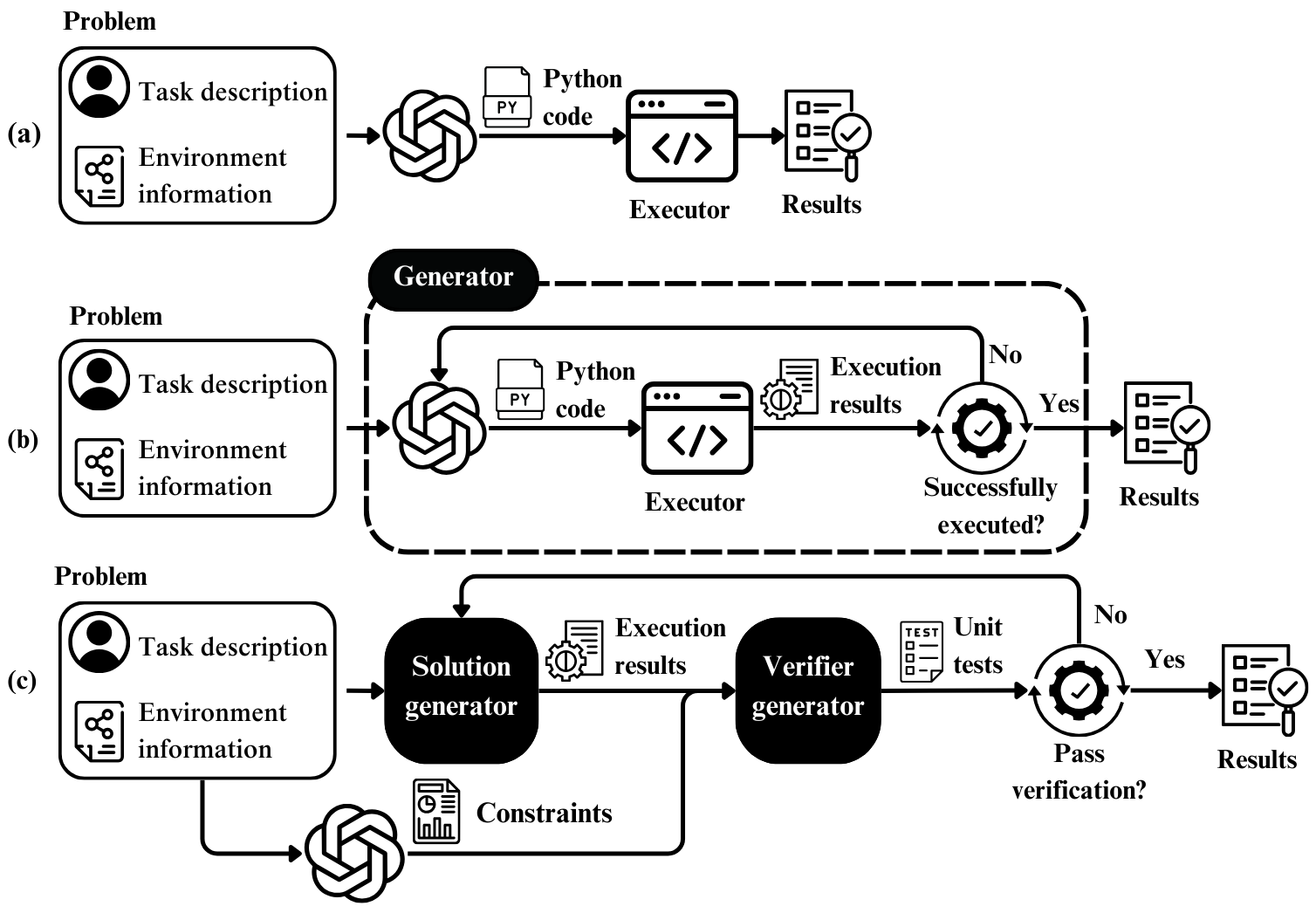}
\caption{Framework overview. (a) \texttt{single attempt}; (b) \texttt{self-debugging}; (c) \texttt{self-debugging with self-verification}. In (c), the solution generator and the verifier generator have the same structure as that in (b).}
\label{fig:framework}
\vspace{-0.22in}
\end{figure*}

\begin{figure*}[ht]
\centering
\vspace{-0.2in}
\includegraphics[width=0.95\linewidth]{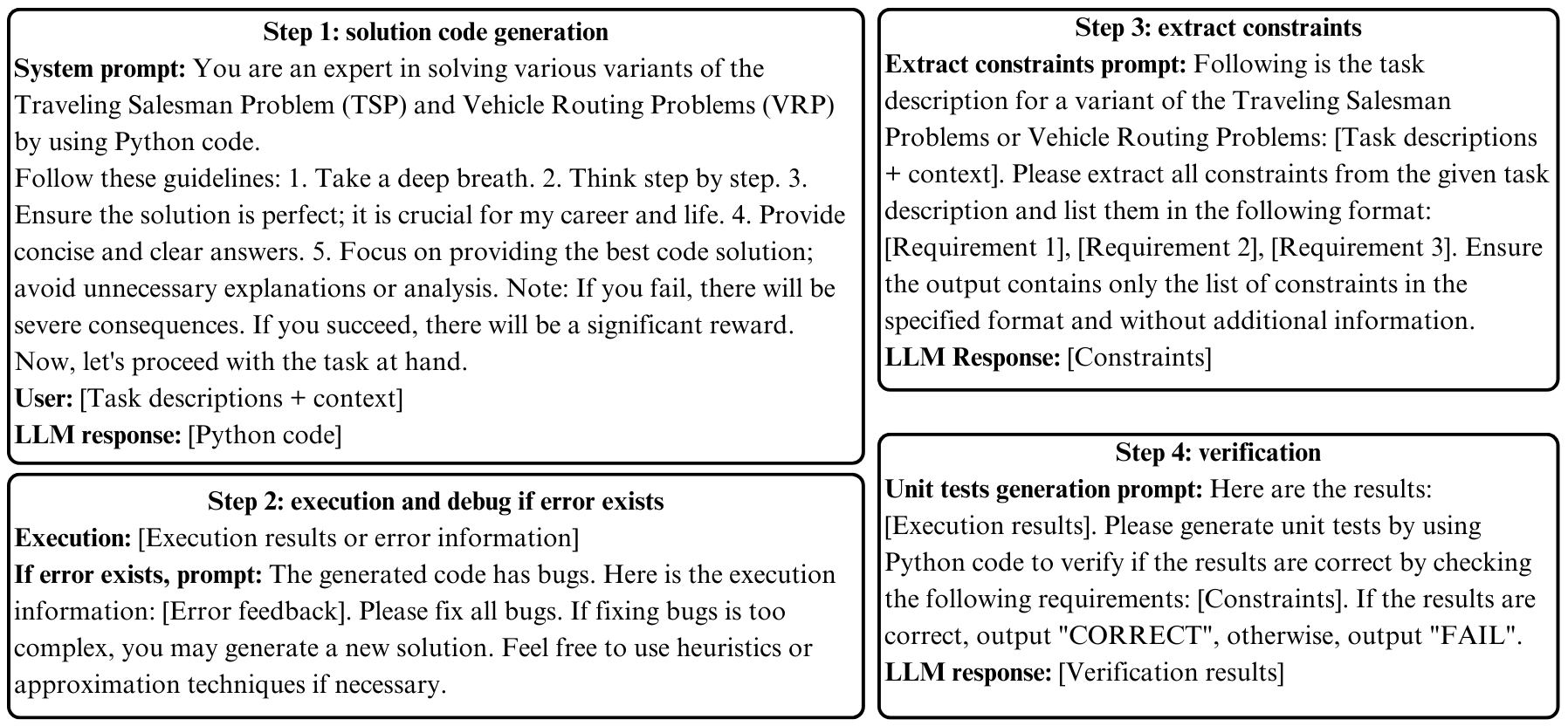}
\caption{Overview of the prompt template.}
\label{fig:prompt_details}
\vspace{-0.24in}
\end{figure*}

Inspired by the observations that LLM-generated solutions might output incorrect results even if the solutions are executable and natural language task descriptions naturally include constraints, we design the third framework by extending the second framework. 
Compared with the second framework, besides asking an LLM to regenerate solutions given execution errors, we ask it to perform two extra operations.
First, after solution execution succeeds and results are acquired, we ask the LLM to extract constraints from natural language task descriptions. 
Second, we ask the LLM to generate unit tests to verify if the acquired results are correct. 
If the results pass unit tests, we assume they are correct. Otherwise, we ask the LLM to regenerate solutions. 
We allow the LLM up to two additional attempts to pass successful execution results for verification. 
In the following, we refer to this framework as \texttt{self-debugging with self-verification}. 
The framework details are shown in Fig.~\ref{fig:framework}. The prompt template is shown in Fig.~\ref{fig:prompt_details}. The \texttt{single-attempt} framework only includes step 1, and the \texttt{self-debugging} framework includes step 1 and step 2. The \texttt{self-debugging with self-verification} framework includes all steps. 



\vspace{-0.1in}
\subsection{LLM Prompt Design: Extra Context}


We start with a prompt containing only task descriptions as shown in Fig.~\ref{fig:overview_v2} as a baseline prompt. Inspired by experts who usually translate natural language task descriptions to exact mathematical formulations, we add exact mathematical formulations of related problems as context based on the baseline prompt. 
We refer to this prompt as \textsc{mathematical formulation} below. 
Inspired by pseudo-code and related research papers that usually help experts in solving routing problems, we designed two different prompts, which use pseudo-code and related research papers as context separately. 
For each task, we select two different pseudo-codes and two different research papers. 
To use a related research paper as context, we ask GPT-4 Turbo to summarize the paper first and use the summary as context. 
We refer to these two prompts as \textsc{pseudo-code} and \textsc{paper} below.

\subsection{Different LLMs}
Our evaluation is based on four LLMs, GPT3.5 (gpt-3.5-turbo-1106), GPT-4 (gpt-4-0613), GPT-4 Turbo (gpt-4-turbo-2024-04-09) and Llama3.1-8B.

\section{Experiment}
In this section, we investigate the ability of LLMs to solve routing problems by answering three main questions: 1. What is the performance of the three different frameworks, 2. What is the performance of LLMs given four different contexts, and 3. What is the performance of the different LLMs? We use four metrics: 

\noindent  \textbf{Success rate}: $\frac{S}{T} \times 100$, where $S$ is the number of attempts that result in a feasible solution, and $T$ is the total number of attempts. 

\noindent  \textbf{Optimality gap}: $\frac{V - V^*}{V^*} \times 100$, where $V$ is the objective value of the solution being evaluated and $V^*$ is the optimal objective value.

\noindent  \textbf{Execution time} is the clock execution time of the generated Python program. 

\noindent  \textbf{Cost} is the total expenses of calling the API of LLMs. 

\subsection{How do the three frameworks perform?}

\subsubsection{The success rate of LLM-generated solutions in different frameworks.}
\begin{figure}[ht]
\centering
\includegraphics[width=0.65\linewidth]{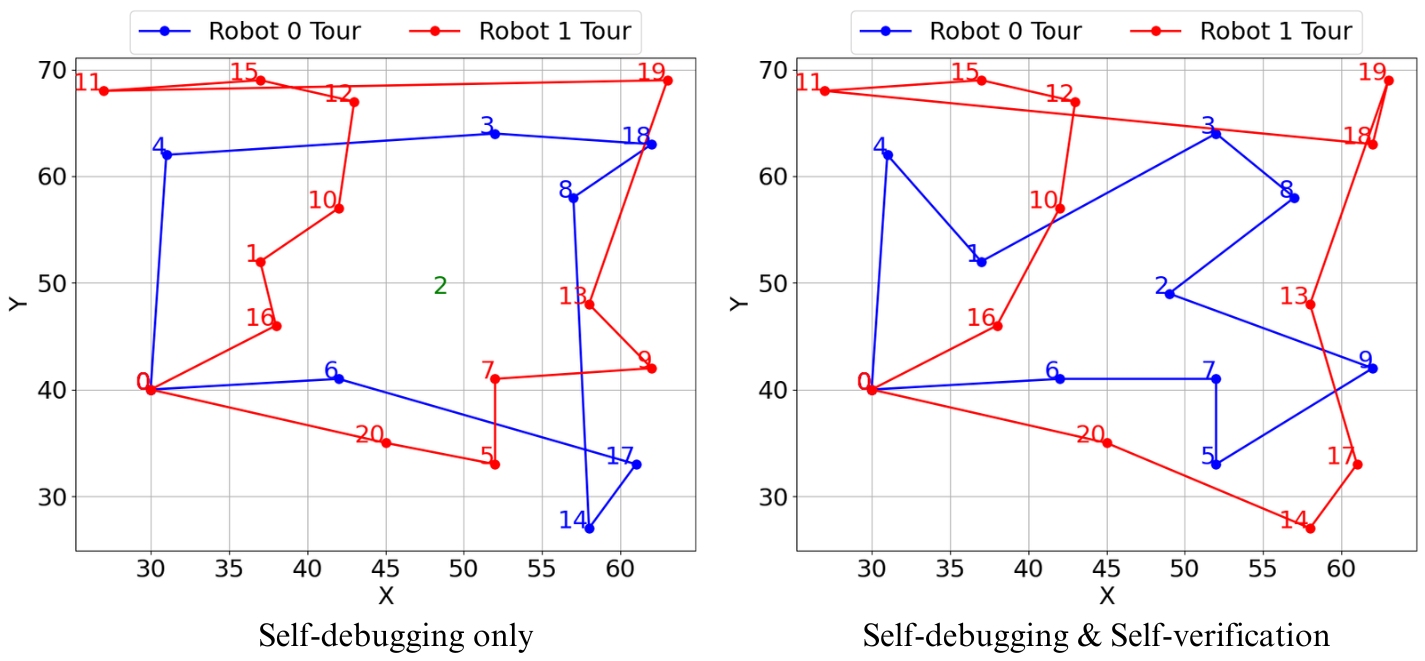}
\vspace{-0.1in}
\caption{CVRP result from \texttt{self-debugging} only framework and that from \texttt{self-debugging with self-verification} framework for P-n21-k2. The results from the \texttt{self-debugging} framework provide an infeasible solution that misses location 2, while by adding self-verification, the framework identifies the incorrect solution and fixes it.}
\label{fig:compare_debug_verify}
\vspace{-0.26in}
\end{figure}

We evaluate the performance of three different frameworks based on GPT-4 Turbo. 
In Table~\ref{tab:eval_framework}, we make several observations: 
First, compared with the baseline, adding \texttt{self-debugging} increases the mean success rate of all task instances by $27.5\%$ (last row in Table~\ref{tab:eval_framework}), which shows that GPT-4 Turbo is able to fix errors in previously generated solutions. 
Second, in addition to \texttt{self-debugging}, incorporating \texttt{self-verification} results in a further $16.0\%$ increase in the mean success rate for GPT-4 Turbo. An intuitive example is shown in Fig. \ref{fig:compare_debug_verify}. 
Third, however, in some tasks, such as GTSP and MinMax $m$-TSP, the framework with both \texttt{self-debugging} and \texttt{self-verification} has a lower success rate compared to the framework with only \texttt{self-debugging}. 
We find that this is because the verifier sometimes misclassifies the correct results as incorrect for some test variants and triggers LLMs to regenerate solutions. When such false alarm feedback is fed into the regenerate process, LLMs are more likely to output inexecutable codes and such codes are less likely to be corrected even with the \texttt{self-debugging} mechanism. Such a phenomenon echoes the mainstream observation of  LLMs' sensitivity to prompts. 

\begin{table}[ht]
\vspace{-0.2in}
    \centering
    \begin{adjustbox}{width=0.95\textwidth}
    \begin{tabular}{llccc|ccc|ccc|ccc}
        \toprule
        \multirow{2}{*}{} & \multirow{2}{*}{Tasks} & \multicolumn{3}{c|}{Avg. Success Rate ($\%$) $\uparrow$} & \multicolumn{3}{c|}{Avg. Opt. Gap ($\%$) $\downarrow$} & \multicolumn{3}{c|}{Avg. Execution Time (s) $\downarrow$} & \multicolumn{3}{c}{Avg. Cost (\$) $\downarrow$} \\
        \cmidrule(lr){3-5} \cmidrule(lr){6-8} \cmidrule(lr){9-11} \cmidrule(lr){12-14}
         &  & Single attempt & Debug & Debug $\&$ Verify  & Single attempt & Debug & Debug $\&$ Verify & Single attempt & Debug & Debug $\&$ Verify & Single attempt & Debug & Debug $\&$ Verify \\
        \midrule
        \multirow{4}{*}{Single} & TSP & 29.33 & 66.67 & 92.00 & 8.69 & 10.66 & 8.94 & 0.23 & 0.19 & 0.28 & 0.02 & 0.05 & 0.15 \\
        & BTSP & 18.67 & 41.33 & 76.00 & 47.77 & 46.68 & 46.58 & 0.43 & 0.40 & 0.24 & 0.03 & 0.05 & 0.23 \\
        & GTSP & 62.67 & 100.00 & 97.33 & 15.91 & 16.37 & 15.77 & 0.10 & 0.08 & 0.07 & 0.02 & 0.04 & 0.13 \\
        & $k$-TSP & 41.33 & 68.00 & 74.67 & 22.70 & 16.71 & 12.77 & 3.69 & 3.13 & 2.37 & 0.02 & 0.04 & 0.14 \\
        \midrule
        \multirow{4}{*}{Multiple} & $m$-TSP & 24.00 & 48.00 & 72.00 & 56.44 & 59.63 & 47.87 & 0.19 & 0.17 & 0.24 & 0.03 & 0.07 & 0.25 \\
        & MinMax $m$-TSP & 40.00 & 68.00 & 64.00 & 23.15 & 32.45 & 46.11 & 1.18 & 0.80 & 0.49 & 0.03 & 0.05 & 0.19 \\
        & MD $m$-TSP & 0.00 & 20.00 & 32.00 & - & 52.40 & 71.06 & - & 1.30 & 0.10 & - & 0.08 & 0.32 \\
        & CVRP & 8.00 & 32.00 & 64.00 & 66.72 & 62.47 & 60.07 & 0.04 & 0.07 & 0.05 & 0.03 & 0.08 & 0.28 \\
        \midrule
        \multicolumn{2}{c}{\textbf{Mean}} & 28.00 & 55.50 & 71.50 & 34.48 & 37.17 & 38.65 & 0.84 & 0.77 & 0.48 & 0.03 & 0.06 & 0.21 \\
        \bottomrule
    \end{tabular}
    \end{adjustbox}
    \vspace{0.08in}
    \caption{Evaluating three different frameworks, \texttt{single attempt}, \texttt{self-debugging}, and \texttt{self-debugging with self-verification}. Each single-robot routing task is evaluated on instances with $10$, $15$, and $20$ locations, and each location number corresponds to five instances. Each multiple-robot routing task is evaluated on five instances with different location number and robot number. All instances in the single-robot tasks and the multiple-robot tasks are evaluated across five runs.}
    \label{tab:eval_framework}
    \vspace{-0.35in}
\end{table}

\subsubsection{The effectiveness of LLM-generated verifier.} 
 We want to quantify the performance of the verifier by measuring two metrics: false negative and false positive. A false negative occurs when the verifier misclassifies correct solutions as incorrect. 
A false positive occurs when the verifier misclassifies incorrect solutions as correct. We collect all the false negative and false positive cases for each task. 
Table~\ref{tab:verifier_misclassify} shows that the false negative rate is 15.00\% (6 / 40) which suggests that the LLM-generated verifier can identify the infeasible solution in most cases and can help to increase the success rate of the \texttt{self-debugging with self-verification} framework. Meanwhile, it should be noted that the false negative rate of the verifier is 30.12\% (75 / 249). Each false negative will give misleading feedback and may cause the success rate to drop. This can explain the exceptions in success rate for some problem instances (e.g., GTSP and MinMax $m$-TSP).

\begin{table}[htbp!]
    \vspace{-0.2in}
    \centering
    \begin{adjustbox}{width=0.46\textwidth}
    \begin{tabular}{llcc}
        \toprule
        & \textbf{Task} & \textbf{False Negative} & \textbf{False Positive} \\
        \cmidrule{1-4}
        \multirow{4}{*}{\textbf{Single}} & TSP & 14 / 50 & 0 / 1 \\
        & BTSP & 14 / 31 & 0 / 5 \\
        & GTSP & 16 / 75 & 0 / 0 \\
        & $k$-TSP & 18 / 51 & 1 / 6 \\
        \midrule
        \multirow{4}{*}{\textbf{Multiple}} & $m$-TSP & 4 / 12 & 0 / 5 \\
        & MinMax $m$-TSP & 6 / 17 & 1 / 5 \\
        & MD $m$-TSP & 3 / 5 & 4 / 11 \\
        & CVRP & 0 / 8 & 0 / 7 \\
        \midrule
        \multicolumn{2}{c}{\textbf{Total}} & 75 / 249 & 6 / 40 \\
        \bottomrule
    \end{tabular}
    \end{adjustbox}
    \vspace{0.1in}
    \caption{The performance of LLM-generated verifier. False negative means the verifier misclassifies correct solutions to incorrect solutions. False positive means the verifier misclassifies the incorrect solutions to correct solutions. For each value, the left side is the number misclassified, and the right side is the total number that the verifier checked.}
    \label{tab:verifier_misclassify}
    \vspace{-0.6in}
\end{table}

\vspace{-0.06in}
\subsubsection{The optimality gap of LLM-generated solutions in different frameworks.}
Table~\ref{tab:eval_framework} shows that neither \texttt{self-debugging} nor \texttt{self-debugging with self-verification} show consistent advantages in decreasing optimality gap even though they show better optimality gap than the \texttt{single attempt} in some tasks, such as $k$-TSP or CVRP. This is likely due to that LLMs use self-designed heuristics to solve the problem and these heuristics do not have an optimality guarantee. Furthermore, the average optimality gap across all tasks indicates that the \texttt{single attempt} outperforms both \texttt{self-debugging} and \texttt{self-debugging with self-verification}, which suggests that the regeneration process increases the success rate at the price of optimality.

\vspace{-0.2in}
\subsubsection{The execution time and cost of LLM-generated solutions in different frameworks.}
Table~\ref{tab:eval_framework} shows that both \texttt{self-debugging} and \texttt{self-debugging with self-verification} have shorter execution time than \texttt{single attempt}. We hypothesize that this may be because the LLM prefers to generate exact algorithms at the first trial, which has high time complexity.
We collect all successfully executed programs from \texttt{single attempt}, \texttt{self-debugging} only, and \texttt{self-debugging with self-verification}, and analyze the ratio of exact algorithms. 
We find that \texttt{single attempt}, \texttt{self-debugging}, and 
\texttt{self-debugging with self-verification} are associated with $51.5\%$, $47.4\%$, and $38.9\%$ exact algorithms respectively. 
This can also partially explain why the framework \texttt{self-debugging with self-verification} has a shorter execution time than the other two frameworks.  
Besides, we find the cost of \texttt{self-debugging} only is twice that of the \texttt{single attempt}, and the cost of \texttt{self-debugging with self-verification} is seven times that of the \texttt{single attempt}.

\vspace{-0.15in}
\subsubsection{The success rate and optimality gap of LLM-generated solutions w.r.t. the number of locations and the number of robots.}

\begin{figure}[ht]
\centering
\vspace{-0.2in}
\includegraphics[width=0.95\linewidth]{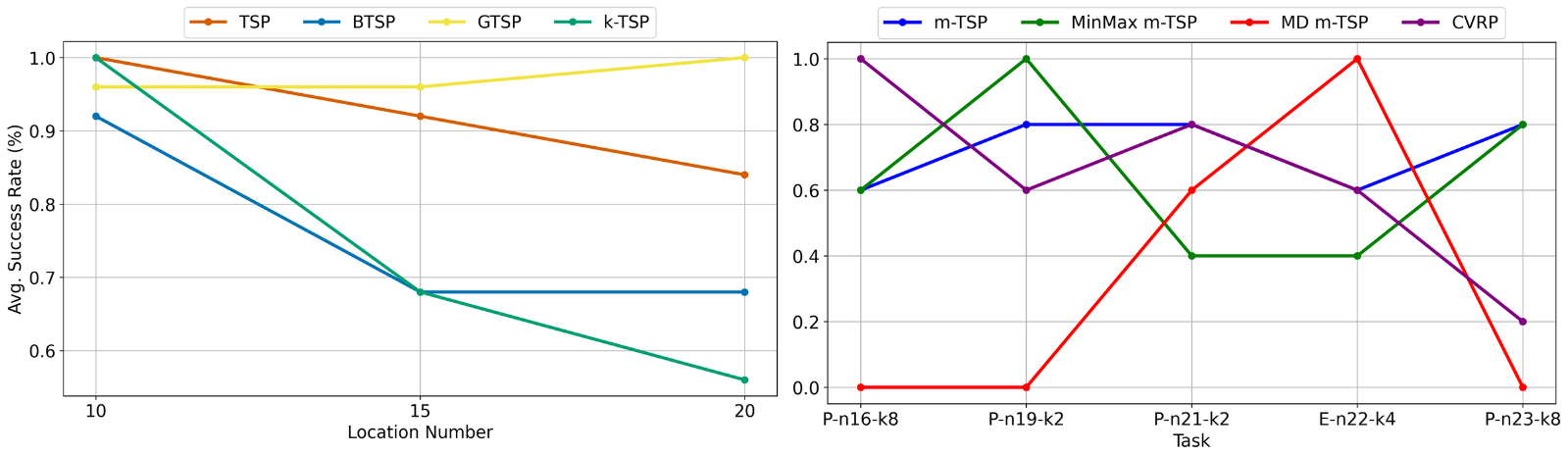}
\vspace{-0.08in}
\caption{GPT-4 Turbo: success rate across different location numbers based on \texttt{self-debugging with self-verification} framework.}
\label{fig:sm_success_loc_num}
\vspace{-0.1in}
\end{figure}

We evaluate the success rate of LLMs w.r.t. different numbers of locations and robots based on the \texttt{self-debugging with self-verification} framework. 
In Figure~\ref{fig:sm_success_loc_num}, we find that for different tasks, GPT-4 Turbo shows different sensitivity to different numbers of locations or robots. Specifically, for TSP, BTSP, $k$-TSP, and CVRP, with the increasing number of locations or the number of robots, the success rate drops. For the rest of the tasks, changes in the number of locations or robots do not show consistent changes in the success rate. Such a phenomenon suggests that LLMs do not have a consistent understanding of the problem but will improvise to prompts each time even though we only change the parameters.

\begin{figure}[ht]
\centering
\includegraphics[width=0.95\linewidth]{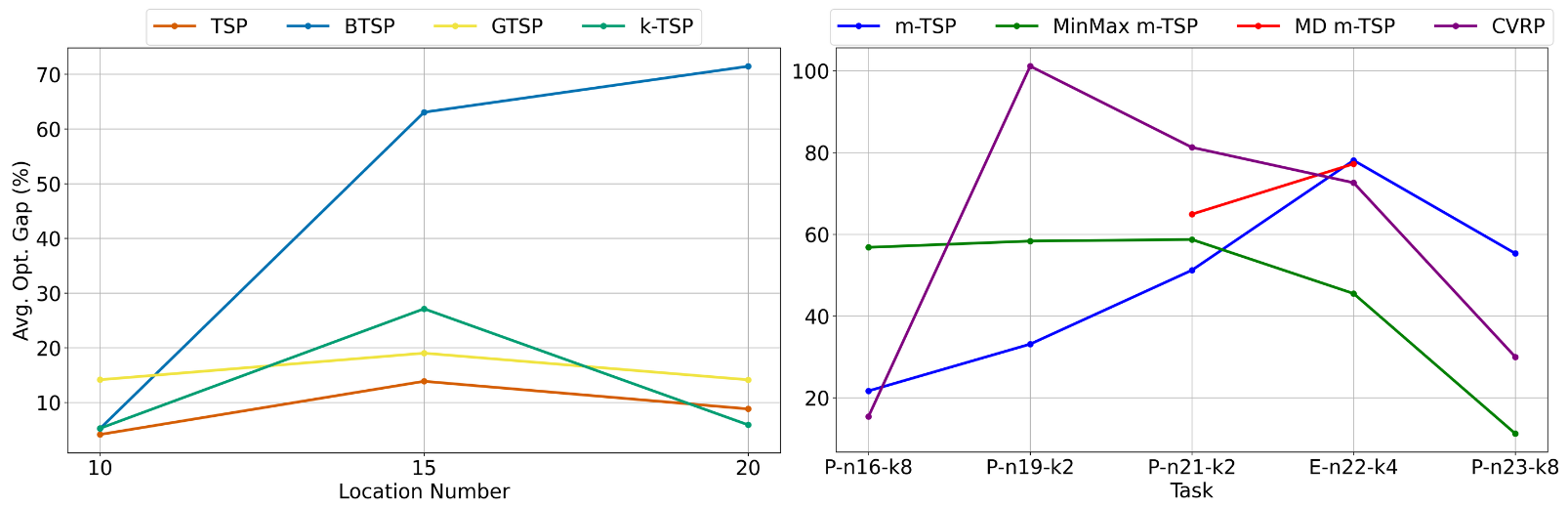}
\caption{GPT-4 Turbo: optimality gap across different location numbers based on \texttt{self-debugging with self-verification} framework.}
\label{fig:sm_opt_gap_loc_num}
\vspace{-0.32in}
\end{figure}

\vspace{-0.1in}
We also evaluate the optimality gap of GPT-4 Turbo w.r.t. a different number of locations and robots based on the \texttt{self-debugging with self-verification} framework. We find (Fig.~\ref{fig:sm_opt_gap_loc_num}) that for different tasks, GPT-4 Turbo shows different sensitivity to different numbers of locations or robots. Specifically, BTSP is sensitive to the number of locations, and CVRP is sensitive to the number of locations but not sensitive to the number of robots. For the rest of the tasks, changes in the number of locations or robots do not show consistent changes in the optimality gap.

\vspace{-0.15in}
\subsubsection{What algorithms do GPT-4 Turbo generate?}
Based on all feasible programs generated by GPT-4 Turbo, we analyzed the algorithms GPT-4 Turbo implemented. 
We classify all algorithms into three categories: exact algorithms, heuristic algorithms, and approximation algorithms.
We find that $38.9\%$ of the generated algorithms are exact and $60.1\%$ are heuristics. Approximation algorithms only take $1.0\%$. 


\vspace{-0.15in}
\subsection{What is the performance of LLMs given four different contexts?}
\begin{figure*}[t!]
\vspace{-0.12in}
    \centering
    \begin{subfigure}[t]{0.5\textwidth}
        \centering
        \includegraphics[height=1.25in]{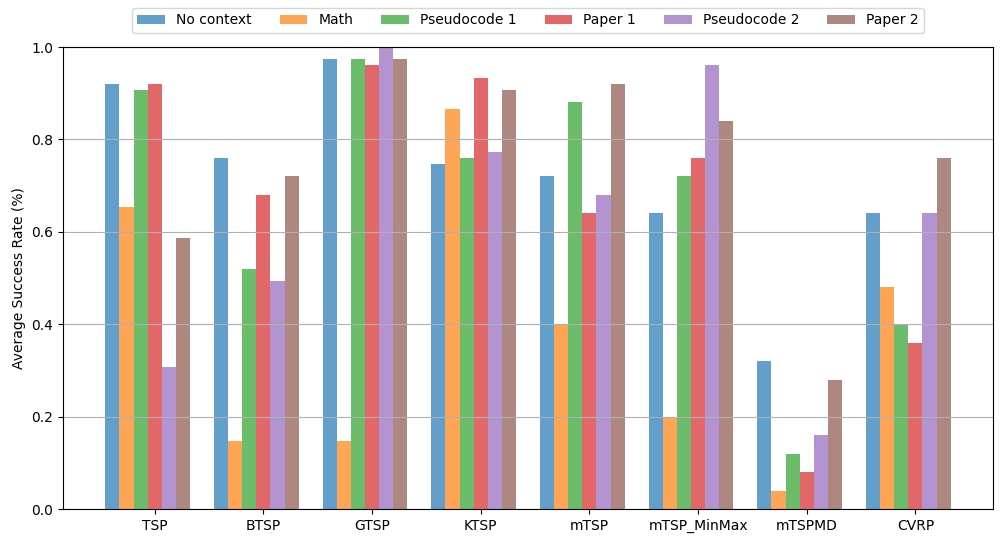}
    \end{subfigure}%
    ~ 
    \begin{subfigure}[t]{0.5\textwidth}
        \centering
        \includegraphics[height=1.25in]{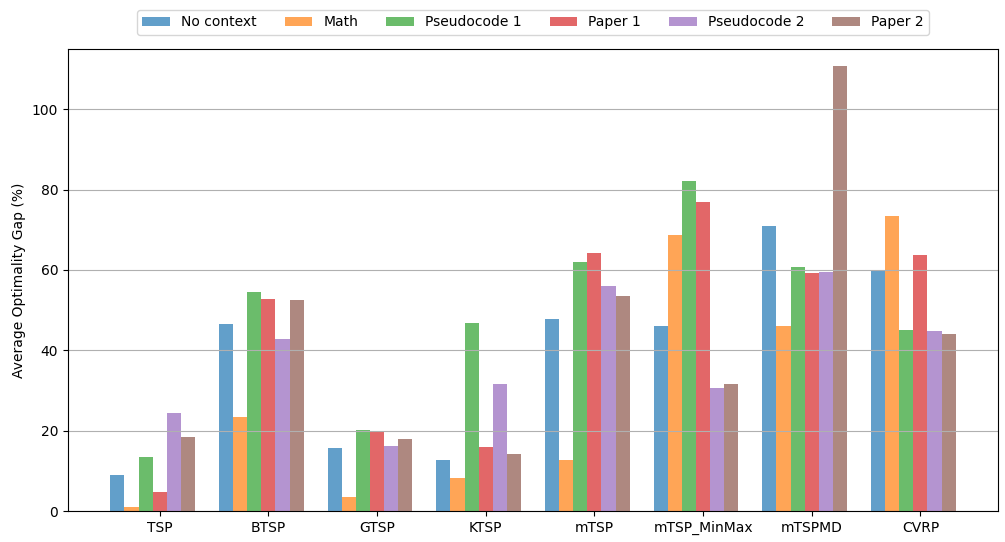}
    \end{subfigure}
    \vspace{-0.26in}
    \caption{GPT-4 Turbo: average success rate (left) and average optimality gap (right) given different contexts based on the \textit{self-debugging with self-verification} framework. Pesudo-code and papers are different across different tasks.}
    \label{fig:debug_verify_success_opt}
    \vspace{-0.2in}
\end{figure*}

Based on the \texttt{self-debugging and self-verification} framework, which performs the best across three different frameworks in most cases, we evaluate the performance of GPT-4 Turbo by adding additional information as context, such as \textsc{mathematical formulation}, \textsc{pseudo-code}, and \textsc{paper} from the published literature. We make the following observations (Fig.~\ref{fig:debug_verify_success_opt}). Adding exact mathematical formulation decreases the success rate and optimality gap of GPT-4 Turbo in most tasks at the same time. 
We find that this may be because the exact mathematical formulation would make GPT-4 Turbo use external libraries, such as Gurobi, PuLP, OR-Tools, and MIP.  GPT-4 Turbo is likely to generate inexecutable codes using these libraries.
We collect the information on the imported libraries of all solutions and calculate the ratio of the solutions that use the above-mentioned libraries. We find that compared with the case without context (Table~\ref{tab:optimization_ratios}), with exact \textsc{mathematical formulations}, the ratio of solutions that use external libraries increases from $2.79\%$ to $41.53\%$, while only $0.62\%$ of the feasible solutions are generated by adding exact \textsc{mathematical formulations} as context. 
When using \textsc{pseudo-code} or related research \textsc{papers} as context, the ratio of using external libraries decreases to less than $0.25\%$.
This suggests that the ability of GPT-4 Turbo to use external libraries is limited. Although adding some \textsc{pseudo-code} or related research \textsc{papers} as context increases the success rate and decreases the optimality gap of GPT-4 Turbo, its performance is sensitive to different \textsc{pseudo-code} and research \textsc{papers}. This shows that its ability to follow instructions given \textsc{pseudo-code} or research \textsc{papers} is also limited.

\begin{table}[ht!]
    \centering
    \vspace{-0.2in}
    \begin{adjustbox}{width=0.95\textwidth}
    \begin{tabular}{lrrrrrrr}
        \toprule
        Category & Gurobi (\%) & PuLP (\%) & OR-Tools (\%) & Pyomo (\%) & MIP (\%) & Docplex (\%) & Total (\%) \\
        \midrule
        No Context & 0.00 & 0.12 & 2.58 & 0.00 & 0.09 & 0.00 & 2.79 \\
        Math & 0.51 & 33.46 & 4.96 & 1.01 & 1.45 & 0.13 & 41.52 \\
        Pseudo-Code & 0.00 & 0.00 & 0.21 & 0.00 & 0.00 & 0.00 & 0.21 \\
        Paper & 0.00 & 0.00 & 0.16 & 0.00 & 0.00 & 0.00 & 0.16 \\
        \bottomrule
    \end{tabular}
    \end{adjustbox}
    \vspace{0.08in}
    \caption{Results on using optimization libraries in LLM-generated programs}
    \label{tab:optimization_ratios}
    \vspace{-0.62in}
\end{table}

\vspace{-0.1in}
\subsection{How does performance vary if we change the LLM?}
\vspace{-0.1in}
\subsubsection{Compare with an open-sourced lightweight LLM using context information}
\begin{figure*}[t!]
\vspace{0.1in}
    \centering
    \begin{subfigure}[t]{0.5\textwidth}
        \centering
        \includegraphics[height=1.25in]{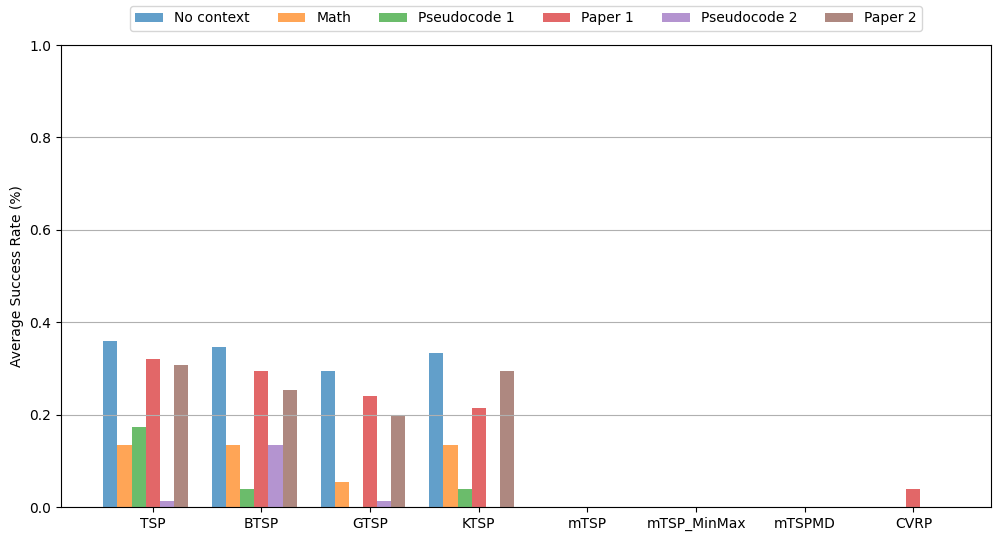}
    \end{subfigure}%
    ~ 
    \begin{subfigure}[t]{0.5\textwidth}
        \centering
        \includegraphics[height=1.25in]{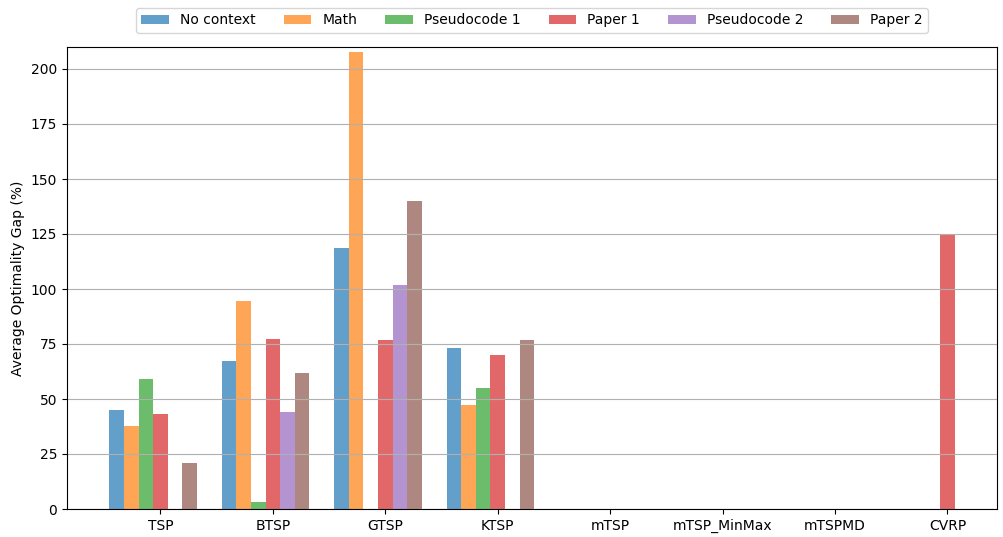}
    \end{subfigure}
        \vspace{-0.1in}
    \caption{Llama3.1-8B: Average success rate (left) and optimality gap (right) given additional information based on the \texttt{self-debugging with the self-verification} framework. \textsc{Pseudo-code} and \textsc{papers} are different across different tasks.}
    \label{fig:llama3_1_debugger_verifier}
    \vspace{-0.24in}
\end{figure*}

We evaluated the performance of an open-sourced lightweight LLM, Llama3.1-8B. In Fig.~\ref{fig:llama3_1_debugger_verifier}, we make two main observations. First, compared with GPT-4 Turbo, the success rate of Llama 3.1-8B across all tasks significantly drops. Besides, Llama3.1-8B basically can not solve multi-robot tasks. Second, unlike GPT-4 Turbo, in Llama 3.1-8B, adding additional information as context does not result in a higher success rate across all tasks. Similar to GPT-4 Turbo, some additional information as context does not consistently decrease the optimality gap of Llama3.1-8B. 

\begin{table}[ht]
    \vspace{0.1in}
    \centering
    \begin{adjustbox}{width=0.96\textwidth}
    \begin{tabular}{l| l|c c c c|c c c c}
        \toprule
        \multirow{2}{*}{} & \multirow{2}{*}{Tasks} & \multicolumn{4}{c|}{Avg. Success Rate ($\%$) $\uparrow$} & \multicolumn{4}{c}{Avg. Opt. Gap ($\%$) $\downarrow$} \\
        \cmidrule(lr){3-6} \cmidrule(lr){7-10}
         &  & Llama3.1 & GPT-3.5 & GPT-4 & GPT-4 Turbo & Llama 3.1 & GPT-3.5 & GPT-4 & GPT-4-Turbo \\
        \midrule
        \multirow{4}{*}{Single} & TSP & 36.00 & 37.33 & 57.33 & 92.00 & 44.90 & 62.76 & 35.06 & 8.94 \\
        & BTSP & 34.67 & 45.33 & 56.00 & 76.00 & 67.41 & 45.04 & 44.68 & 46.58 \\
        & GTSP & 29.33 & 60.00 & 85.33 & 97.33 & 118.76 & 131.86 & 13.15 & 15.77 \\
        & $k$-TSP & 33.33 & 36.00 & 78.66 & 74.67 & 73.09 & 5.46 & 30.83 & 12.77 \\
        \midrule
        \multirow{4}{*}{Multiple} & $m$-TSP & 0.0 & 4.00 & 12.00 & 72.00 & - & 40.59 & 73.58 & 47.87 \\
        & MinMax $m$-TSP & 0.0 & 0.0 & 16.00 & 64.00 & - & - & 14.59 & 46.11 \\
        & MD $m$-TSP & 0.0 & 0.0 & 0.0 & 32.00 & - & - & - & 71.06 \\
        & CVRP & 0.0 & 0.0 & 8.00 & 64.00 & - & - & 32.87 & 60.07 \\
        \midrule
        \multicolumn{2}{c|}{\textbf{Mean}} & 16.67 & 22.83 & 39.17 & 71.50 & 76.04 & 57.14 & 34.97 & 38.65 \\
        \bottomrule
    \end{tabular}
    \end{adjustbox}
    \vspace{0.1in}
    \caption{Based on the \texttt{self-debugging with self-verification} framework, we evaluate $4$ different LLMs: Llama 3.1-8B, GPT-3.5, GPT-4, and GPT-4-Turbo. Each single-robot task is evaluated on $10$, $15$, and $20$ locations, each location number with $5$ instances. Each multiple-robot task is evaluated on $5$ instances with different location numbers and robot numbers. All instances in the single-robot tasks and the multiple-robot tasks are evaluated across $5$ runs.}
    \label{tab:diff_llms}
    \vspace{-0.36in}
\end{table}

\vspace{-0.2in}
\subsubsection{Compare with four different LLMs without context information}
Based on the \texttt{self-debugging with self-verification} framework, we evaluate four different LLMs, Llama 3.1-8B, GPT-3.5, GPT-4, and GPT-4-Turbo on success rate and optimality gap across all variants of tasks. 
In Table~\ref{tab:diff_llms}, we make several observations. 
First, although Llama 3.1-8B performs the worst, the gap between Llama 3.1-8B and GPT-3.5 is around $6\%$. 
Second,  Llama 3.1-8B, GPT-3.5, and GPT-4 all show a low mean success rate ( $ \leq 12\%$) on multiple-robot routing problems. Compared with these three LLMs, the success rate of GPT-4 Turbo in solving multiple-robot routing problems is significantly higher. 
Third, the mean optimality gap among all GPT LLMs is at a similar level although GPT-4 shows the lowest average optimality gap.

\vspace{-0.15in}
\section{Discussions and Future Work}
Although our work shows that LLMs are useful in solving robot routing in many cases, there is still plenty of room to improve. To increase the ability of LLMs to generate feasible solutions with low optimality gaps or even optimal solutions, we propose two directions.
\vspace{-0.15in}
\subsection{Direction 1: Building on existing LLMs.}
We suggest two avenues to build frameworks to improve on the success rate and optimality gap - one is in optimization and the other is in building a library. 

\textbf{Optimization}
Given natural language task descriptions, we can ask LLMs to generate fitness functions for routing problems. We then use fitness functions to evaluate feasible solutions generated by LLMs.
We continue asking LLMs to generate solutions until the optimality gap does not decrease anymore or the decrease in the optimality gap falls below a threshold between two iterations. 

\textbf{Build a library}
We can collect high-quality code implemented by experts and put that in a library. For each code, we can generate a (task description, code) pair.
Every time a new task arrives, we can retrieve the code with similar task descriptions from the library and use the selected code as context. 
Every time LLMs generate better solutions than the solutions in the library, or LLMs generate feasible solutions for new tasks, we can add the (task description, code) pair into the library.  
\vspace{-0.1in}
\subsection{Direction 2: Fine-tune LLMs}
Since LLMs are trained on diverse data to solve general problems, this might decrease their ability to generate near-optimal solutions or optimal solutions for routing. 
Instead of waiting for LLMs to be mature enough to solve routing problems with near-optimal solutions or even optimal solutions, we can fine-tune an existing open-source LLM by ourselves to improve its ability to solve routing problems. Specifically, we can collect high-quality data from human experts and websites or work with LLMs to generate high-quality data through well-crafted prompts. 

\section{Conclusion}

We investigated the ability of Large Language Models to solve robot routing problems. We start by constructing a dataset with 80 unique problems across 8 variants of routing problems. We then design 3 LLM frameworks and 4 types of context. We find both self-debugging and self-verification benefit LLMs to obtain higher success rates while not helping to decrease the optimality gap. 
Additionally, we find LLMs are sensitive to different contexts. Exact mathematical formulation, when provided as context, decreases the optimality gap but significantly decreases the success rate. 
Pseudo-code and research papers, when provided as context, increase the success rate of LLMs, but do not help with the optimality gap. 
We concluded by proposing two directions that might help increase the performance of LLMs in solving robot routing.

\printbibliography
\end{document}